\documentclass[runningheads]{llncs}
\usepackage[utf8]{inputenc}
\usepackage[usenames, dvipsnames]{color}
\usepackage{todonotes}
\usepackage[normalem]{ulem}
\usepackage{float}
\usepackage{amsmath}
\usepackage{amsfonts}
\usepackage{amssymb}
\usepackage{graphicx}
\usepackage{hyperref}
\usepackage{amsmath}
\usepackage{graphicx}
\usepackage{url}
\usepackage{subfig}
\usepackage{booktabs}
\usepackage{tabularx}
\usepackage{enumerate}
\usepackage[shortlabels]{enumitem}

\newcommand{\furl}[1]{\footnote{\scriptsize \url{#1}}}

\begin{document}
\titlerunning{Completion of Scholarly Knowledge Graph Using TransE with Soft Margins}
\title{
Soft Marginal TransE for \\ Scholarly Knowledge Graph Completion
}

\author{Mojtaba Nayyeri, Sahar Vahdati, Jens Lehmann, Hamed Shariat Yazdi}
\institute{%
University of Bonn, Germany \\
\email{familyname@cs.uni-bonn.de}}
\authorrunning{Nayyeri et al.}

\maketitle     
\setcounter{footnote}{0}
%
\begin{abstract}
Knowledge graphs (KGs), i.e.\ representation of information as a semantic graph, provide a significant test bed for many tasks including question answering, recommendation, and link prediction. 
Various amount of scholarly metadata have been made available as knowledge graphs from the diversity of data providers and agents.
However, these high-quantities of data remain far from quality criteria in terms of completeness while growing at a rapid pace. 
Most of the attempts in completing such KGs are following traditional data digitization, harvesting and collaborative curation approaches.
Whereas, advanced AI-related approaches such as embedding models - specifically designed for such tasks - are usually evaluated for standard benchmarks such as Freebase and Wordnet.
The tailored nature of such datasets prevents those approaches to shed the lights on more accurate discoveries. 
Application of such models on domain-specific KGs takes advantage of enriched metadata and provides accurate results where the underlying domain can enormously benefit. 
In this work, the TransE embedding model is reconciled for a specific link prediction task on scholarly metadata. 
The results show a significant shift in the accuracy and performance evaluation of the model on a dataset with scholarly metadata. 
The newly proposed version of TransE obtains 99.9\% for link prediction task while original TransE gets 95\%. 
In terms of accuracy and Hit@10, TransE outperforms other embedding models such as ComplEx, TransH and TransR experimented over scholarly knowledge graphs. 

\end{abstract}

%
        
        
    
    
%
\section{Introduction}
\label{sec:intro}
With the rapid growth of digital publishing, researchers are increasingly exposed to an incredible amount of scholarly artifacts and their metadata.
The complexity of science in its nature is transparent in such heterogeneous information which generates a baseline for data analytic approaches inheriting big data characteristics. 
In the past decade, the importance of this topic and intangible interest on analyzing scholarly metadata from different research fields such as natural, computational and social science has emerged the ``Science of Science'' concept (SciSci)~\cite{fortunato2018science}. 
To facilitate scholarly communication and management of such metadata, powerful technologies are being developed in order to capture the semantic relations and entities within research artifacts across disciplines.
The ultimate objective of such attempts ranges from providing services for researches to measuring research impact and accelerating science.
Recommending potential joint work as co-authors, suggesting relevant events, relevant papers are among the most desirable services in the scholarly domain~\cite{xia2017big}.

So far most of the methods in the aforementioned scenarios, use semantic similarity and graph clustering techniques to do recommendations~\cite{ammar2018construction}. 
Such approaches are restricted by the available information in the user/item profiles and are incapable of encoding other potentially useful information.
Approaches for knowledge extraction from huge networks by uncovering patterns and predicting emergent properties of the network can facilitate link prediction activities. 

Knowledge graphs (KG) are proven to be quite useful in modeling and representing factual information and have shown to be quite useful in various domains \cite{arber2015comparative}.
In this regard KG creation and integration techniques are being used in research infrastructures in order to allow intelligent services and deduce insightful information.
Formally, a KG is a collection of triples and a triple is fact of the form $(h,r,t)$, where $h$ and $t$ are respectively called head and tail entities and $r$ is the relation between them, e.g.\ (Albert Einstein, co-author, Boris Podolsky).
Despite broad usages and applications of KGs, the existing KGs remain far from complete and automatically completing them, i.e.\ inferring new facts with a high confidence value of truth, has gained much attention so far. 

As expected, Scholarly Knowledge Graphs (SKGs) are also incomplete, mainly due to the document-oriented workflow of scientific publishing and communication as well as the practiced data harvesting and integration approaches~\cite{tharani2015linked,schirrwagen2013data,wan2019aminer}.
Despite the importance of the scholarly domain, automatic KG completion methods using link discovery tools have rarely been studied~\cite{henck2019} as the graphs are heterogeneous and complex by nature~\cite{xia2017big}.
Therefore, SKG completion is a challenge that should be properly addressed in order to help the community be benefited in their scientific activities.

So far different methods have been introduced for KG completion as a generic problem \cite{Nickel2015ReviewRelationalMLforKG}.
The most recent successful methods try to capture the semantic and structural properties of a KG by encoding the information as multi-dimensional vectors.
Such methods are known as knowledge graph embedding (KGE) models in the literature \cite{wang2017surveyEmbeddings}. 
The core idea behind KGE is that symbolic entities and their associated relationships in a KG can be properly represented by global latent features reflected in numerical vectors called embeddings. 

TransE \cite{bordes2013translating} is one of the mostly used embedding model  and is the motivating base model for other models derived form it like 
TransH, TransR \cite{TransH,TransR} etc.
Other kinds of embedding models also exist such as DistMult~\cite{Distmult} and DKRL~\cite{xie2016representation}.
To the best of our knowledge, except the previously highlighted models~\cite{henck2019}, other knowledge graph embedding models have not been investigated on the domain of scholarly knowledge graphs.
In~\cite{henck2019}, authors reported TransE to be the best of the five mentioned, reaching to 50.72\% on the filtered Hit@10 and 647.42 on the filtered mean rank.
The authors used PyKEEN~\cite{ali2018biokeen} as the implementation framework.

TransE has a margin ranking loss, meaning that the model tries to rank true triples lower (more relevant) than the false triples.
In the learning phase, this is achieved by enforcing a margin between the positive and negative examples, forcing true triples are ranked lower by at least the size of the given margin.
Since the margin is rigidly set, the model is not flexible enough
to deal with cases that a triple is wrongly labeled as true negative, resulting a poorer performance.

In this work, we propose a modified version of the Margin Ranking Loss (MRL) to train TransE model. We name the model as TransESM\footnote{TransESM: TransE with Soft Margins.}, which considers margins as soft boundaries in its optimization.
Soft margins allow false negative samples be slightly slide into the margin, mitigating the adverse effects of false negative samples.
In contrast to \cite{henck2019}, we did not use the PyKEEN implementation, rather we re-implemented the TransE model as well as TransESM and trans-RS \cite{zhou2017learning}.
In our implementation, we used AdaGrad with the settings reported in~\cite{ComplEx}.
We got 91\% as filtered Hit@10 for our implementation of the TransE model on the scholarly data set.
Our experiments showed that adding soft margins could improve Hit@10 by 8\%, reaching to 99\% filtered Hit@10 for our new TransESM model.

The remaining part of this paper proceeds as follow.
Section \ref{sec:kg} represents details of the scholarly knowledge graph that is created for the purpose of applying link discovery tools. 
Section \ref{sec:pre} provides a summary of preliminaries required about the embedding models. 
As the focus embedding model of this paper, TransE, is represented in section \ref{sec:tran}.
Section \ref{sec:development} contains the given approach and description of the changes to the TransE model.
An evaluation of the proposed model on the represented scholarly knowledge graph is shown in section \ref{sec:evaluation}.
In section \ref{sec:conclusion}, we lay out the insights and provide a conjunction of this research work.

\section{Scholarly Knowledge Graphs}
\label{sec:kg}

The proposed extension to the TransE model on the tasks of link prediction and triple classification. 
A scholarly knowledge graph that has been created and used in a previous version of this work \cite{henck2019} is also considered in this work.
Initially, the knowledge graph is created after a systematic analysis of the available scholarly metadata in RDF format on the Web.
This includes DBLP\footnote{\url{https://dblp2.uni-trier.de/}}, Springer Nature SciGraph Explorer\footnote{\url{https://springernature.com/scigraph}}, Semantic Scholar\footnote{\url{https://semanticscholar.org}} and the Global Research Identifier Database (GRID)\footnote{\url{https://www.grid.ac}} with metadata about institutes. 
The primary objective of creating this KG is to provide scientific recommendations such as collaboration potentials using link perdition.

The dataset, used for model training, comprises 45,952 triples which is split into triples of training/validation/test sets. 

Each instance in the scholarly knowledge graph is equipped with a unique ID to enable the identification and association of the KG elements.
The knowledge graph consist of the following core entities:
\begin{itemize}
    \item\noindent\textbf{Papers:} refer to the publications of researchers as part of a scientific event, such as a conference. 
    Besides an \emph{ID}, each paper entity in this knowledge graph is enriched with the properties \emph{title} and \emph{year of publication}.
    \item\noindent\textbf{Conferences:} are events in which the scientific publications are hosted. 
    The metadata corresponding to this entity are the \emph{name} of the conference as well as the \emph{year} in which the event took place. 
    The association of papers and conferences in the KG is represented by the relation \emph{wasPublishedIn}.
    \item\noindent\textbf{Authors:} of scientific publications are included in the KG with the property \emph{name}.
    The papers co-authored by a researcher are associated using the relation \emph{isAuthorOf}. Moreover, the KG includes the relation \emph{publishedIn} which corresponds to a mapping of authors and the conferences they published in.
    \item\noindent\textbf{Departments:} correspond to the affiliations of the authors in the KG. The metadata contains a \emph{label} for each instance of this type which typically includes \emph{name}, \emph{city}, \emph{state}, and \emph{country} of the department or the organization it corresponds to. The relation \emph{isAffiliatedIn} represents associations of an author and a department.
\end{itemize}

\begin{table}[tb]
\centering
\begin{tabular}{lrrrrr}
\toprule
\textbf{Dataset type} & \textbf{Collaboration} & \textbf{Publication} & \textbf{Affiliation} & \textbf{Venue} \textbf{\#total} \\
\midrule
Train Dataset  &  12,711 & 8,670 & 12,143 & 39,952    \\
Validation Dataset & 651  & 438 & 588 & 2,000  \\
Test Dataset & 1,953 & 1,264 & 1,770 & 6,000  \\
\bottomrule
\end{tabular}
\caption{\textbf{Dataset Statistics.} The number of triples that are used in different datasets are shown per each relationship.}
\label{tab:stats}
\end{table}

\section{Preliminaries}
\label{sec:pre}
Let $\mathcal{E}, \mathcal{R}$ be the sets of entities and relations respectively. 
A Knowledge graph can be roughly represented as a set $\mathcal{K} = \{(h,r,t) | h, t \in \mathcal{E}, r \in \mathcal{R}\}$ in which 
$h, t$ refer to the subject and object respectively and $r$ refers to a relation. 

The goal of KGE is to obtain vector representation for entities and relations in a KG. In this paper, vectors of $h,r,t$ are shown in boldface i.e.\ $\textbf{h}, \textbf{r} $, $\textbf{t} \in \mathbb{R}^d$, and $d$ is the dimension of the embedding space. 

Each KGE model defines an score function $f_r(h,t)$. The score function gets a triple ($h,r,t$) and returns a value determining if the triple is a fact or not. 
A KGE model initializes embedding vectors randomly. Then it updates the embedding vectors by optimizing a loss function $\mathcal{L}$.
Since typically many variables should be adjusted in the learning process, Stochastic Gradient Descent (SGD) method is commonly used for the optimization of the loss function.

\section{Review of TransE model}
\label{sec:tran}

Earlier translational KGE models have simpler scoring functions \cite{wang2017surveyEmbeddings}. 
In recent works, more complicated scoring functions have been proposed. 
The intuition behind of using more complicated scoring functions is that they should better capture underlying information in KGs. 
However, it is shown that more complicated model are due to risk of over fitting to data or noise and being incapable of performing well on unseen data \cite{trouillon2017knowledge}.
In this regard, it is reported that TransE \cite{bordes2013translating}, as one of the simplest translation based models, outperformed more complicated KGEs in \cite{henck2019}.

The initial idea of TransE model is to enforce embedding of entities and relation in a positive triple ($h,r,t$) to satisfy the following equality:
 \begin{align}
     \textbf{h} + \textbf{r} = \textbf{t}
 \end{align}
where \textbf{h}, \textbf{r} and \textbf{t} are embedding vectors of head, relation and tail respectively. TransE model defines the following scoring function:
\begin{align}
    f_r(h,t) = \| \textbf{h} + \textbf{r} - \textbf{t} \|
\end{align}

The function takes a triple ($h,r,t$) and returns the value that measures the degree of correctness of the triple. A lower value for the score function indicates that the triple is more plausible comparing to those triples with higher values. 

TransE uses margin ranking loss (MRL) to optimize the embedding vectors of entities and relations. MRL computes embedding of entities and relation in a way that a positive triple gets lower score value than its corresponding negative triple. The least difference value between score of positive and negative samples is margin. The MRL is defined as follows:
\begin{align}
    \mathcal{L} = \sum \sum \,
    [f_r(h,t) + \gamma - f_r(h^{'}, t^{'})]_+
\end{align}
where $[x]_+=\max(0,x)$.

MRL has two disadvantages.
The first disadvantage is that the margin can slide. 
Assume that $f_r(h_1,t_1)=0 $ and $f_r(h^{'}_1,t{'}_1) = \gamma$, or $f_r(h_1,t_1)=\gamma $ and $f_r(h^{'}_1,t{'}_1) = 2 \gamma$ are two possible scores for a triples and its negative sample. 
Both of these scores get minimum value for the optimization causing the model to become vulnerable to a undesirable solution. 
The second disadvantage of MRL is that embeddings are adversely affected by false negative samples. 

To tackle the first problem, \cite{zhou2017learning} revises the MRL by adding a term to limit maximum value of positive score:

\begin{align}
    \mathcal{L}_{RS} = \sum \sum \,
    [f_r(h,t) + \gamma - f_r(h^{'}, t^{'})]_+ + \lambda [f_r(h,t) - \gamma_1]_+
\end{align}
 \cite{zhou2017learning} shows $\mathcal{L}_{RS}$ significantly improves the performance of TransE. Authors in \cite{zhou2017learning} denote TransE which is trained by $\mathcal{L}_{RS}$ as TransE-RS.
 
\section{Soft Marginal TransE Model}
\label{sec:development}

This section proposes a new optimization framework for training TransE model. 
The framework fixes the second problem of MRL and its extension mentioned in the previous section. 
The optimization utilizes slack variables to mitigate negative effect of the generated false negative samples. 
In contrast to margin ranking loss, our optimization uses soft margin.
Therefore, uncertain negative samples are allowed to slide inside of margin.

\begin{figure}[!h]
\centering
\includegraphics[width=1.0\textwidth]{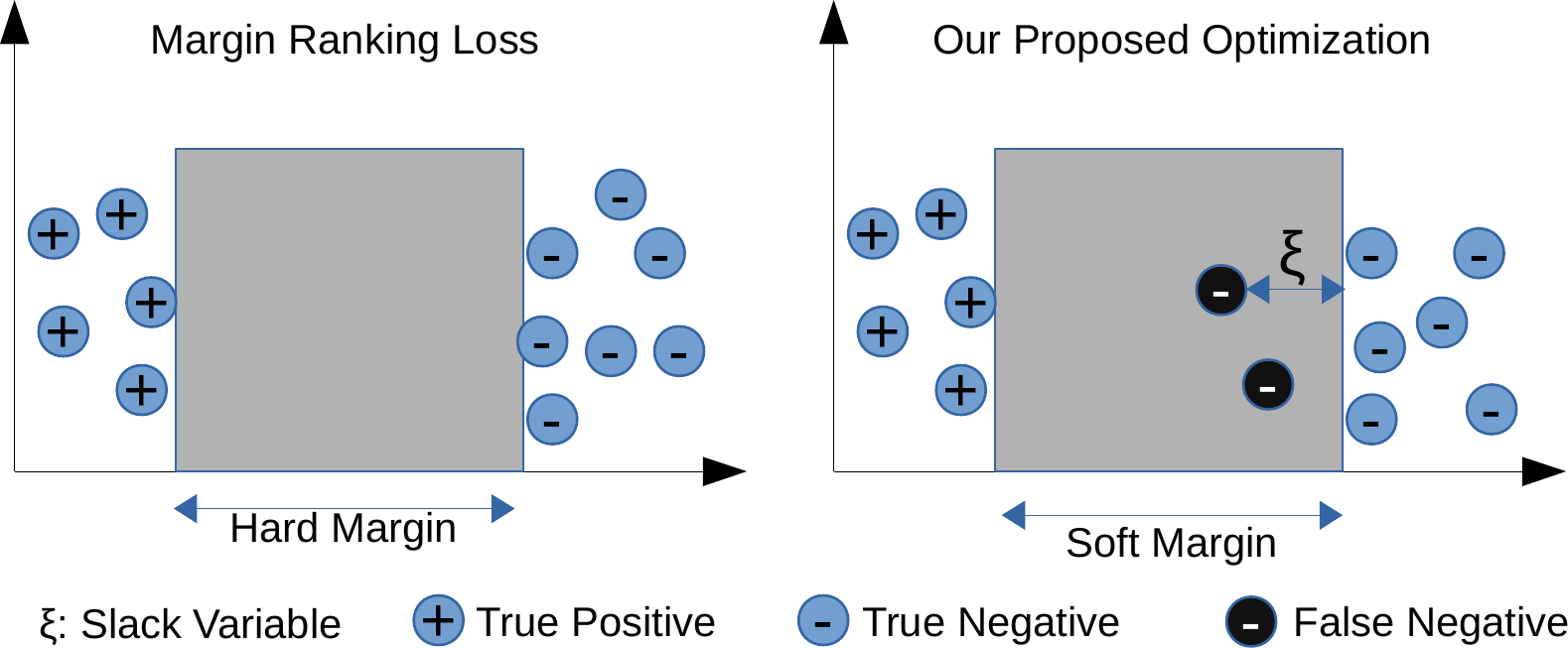} 
\caption{Optimization of margin ranking loss.}
	\label{fig:F1}
\end{figure}

\autoref{fig:F1} visualizes the separation of positive and negative samples using margin ranking loss and our optimization problem. It shows that the proposed optimization problem allows false negative samples to slide inside the margin by using slack variables ($\xi$). In contrast, margin ranking loss doesn't allow false negative samples to slide inside of the margin. Therefore, embedding vectors of entities and relations of false negative samples are adversely affected by false negative samples.  
The mathematical formulation of our optimization problem is as follows:
\begin{align} 
\begin{cases}
\min_{\xi_{h,t}^r} \sum_{(h,r,t) \in S^+}^{~} \; {\xi_{h,t}^r}^2 
\\ 
 \text{s.t.}\\
f_{r}(h,t) \leq \gamma_1 , \; (h,r,t) \in S^+ \\
f_{r}(h^{'}, t^{'}) \geq \gamma_2 - {\xi_{h,t}^r} , \; (h^{'}, r, t^{'}) \in S^-  \\
\xi_{h,t}^r \geq 0
\end{cases}
\label{mainOpt}
\end{align}
where $f_r(h,t) = \| \textbf{h} + \textbf{r} - \textbf{t} \|$ is the score function of TransE, $S^+, S^-$ are positive and negative samples sets. $\gamma_1 \geq 0$ is the upper bound of score of positive samples and $\gamma_2$ is the lower bound of negative samples. $\gamma_2 - \gamma_1$ is margin ($\gamma2 \geq \gamma1$). $\xi_{h,t}^r$ is slack variable for a negative sample that allows it to slide in the margin. $\xi_{h,t}^r$ helps the optimization to better handle uncertainty resulted from negative sampling. 

The objective ($\sum \xi_{h,t}^r$) in the problem \ref{mainOpt} is quadratic. Therefore, it is convex. Moreover, all three constraints can be represented as convex sets. Therefore, the constrained optimization problem \ref{mainOpt} is convex. As a conclusion, it has a unique optimal solution. The optimal solution can be obtained by using different standard methods e.g.\ penalty method. The goal of the problem \ref{mainOpt} is to adjust embedding vectors of entities and relations. Therefore, a lot of variables participate in optimization. In this condition, using batch learning with stochastic gradient descent (SGD) is preferred. In order to use SGD, constrained optimization problem \ref{mainOpt} should be converted to unconstrained optimization problem. The following unconstrained optimization problem is proposed instead of \ref{mainOpt}.

\begin{align} 
\min_{\xi_{h,t}^r} \sum_{(h,r,t) \in S^+} \lambda_0 {\xi_{h,t}^r}^2 + \lambda_1 \max(f_{r}(h,t) - \gamma_1,0) + \lambda_2\, \max(\gamma_2 - f_{r}(h^{'},t^{'}) - {\xi_{h,t}^r},0) 
\label{uncOpt}
\end{align}

The problem \ref{mainOpt} and \ref{uncOpt} may not have the same solution. However, we experimentally see that if $\lambda_1$ and $\lambda_2$ are properly selected, the results would be improved comparing to margin ranking loss.

\section{Evaluation}
\label{sec:evaluation}
This section presents the evaluations of TransSM, our proposed version of TransE, over a scholarly knowledge graph. 
Our model is compared with existing embedding models in terms of Mean Rank (MR) and Hit@10. These metrics have been widely used for evaluation of embedding models on link prediction task. 
A set of pre-processing steps have been done in order to compute Mean Rank as:
\begin{itemize}
    \item head and tail of each test triple is replaced by all entities in the dataset,
    \item scores of the generated triples are computed and sorted,
    \item the average rank of correct test triples is reported as MR. 
\end{itemize}

The computation of Hit@10 is obtained by replacing all entities in the dataset in terms of head and tail of each test triples. 
The result is a sorted list of triples based on their scores.
The average number of triples that are ranked lower than 10 is reported as Hit@10 as represented in table\ref{tab:data}.

Embedding dimension ($d$), upper bound of positive samples ($\gamma_1$), lower bound of negative samples ($\gamma_2$), regularization term ($\lambda_0$) are of hyper-parameters of TransSM. They are searched in the sets $\{50,100,200\}, \{0.1, 0.2, ... , 2\}, \{0.2, 0.3, ... , 2.1\}$ and $\{0.01, 0.1, 0, 1, 10,100\}$ on validation set. We set $\lambda_1 = 1, \lambda_2 = 1$ for simplicity.

The results mentioned in the table \ref{tab:data} validate that TransESM significantly outperformed other embedding models in both MR and Hit@10. 

\begin{table}[tb]
\centering
\begin{tabular}{lcccc}
\toprule
\textbf{} & \textbf{~~~~~~Mean Rank} & & \textbf{~~~~~~Hit@10} & \\
\midrule
\textbf{Setting} & \textbf{Raw} & \textbf{Filtered} & \textbf{Raw} & \textbf{Filtered} \\
\midrule
TransE & 27.4 & 2.17 & 76.6 &  95.0\\
ComplEx & - & - & -  & 56.2\% \\
TransH & 974.88 & 985.16 & 18\% & 21.7\% \\
TransR & 1258.36 & 1200.35 & 23.52\% & 28.09\% \\
TransE-RS & - &  3.03 & - &  95.6\\
\midrule
TransESM & 25.1 & 1.67 & 79.2\% & 99.9\% \\
\bottomrule
\end{tabular}
\caption{\textbf{Experimental Results.} Results of TransE and TransESM are based on our code. For ComplEx we ran the code of authors. Results of TransH and TransR are taken from \cite{henck2019}}
\label{tab:data}
\end{table}

\autoref{tab:rank2} shows the number of recommendations and their rank among these 50 top prediction for all of the 9 sample researchers. 
\begin{table}[h]
\centering
\begin{tabular}{lcr}
\toprule
\textbf{Author} & \textbf{\#Recom.} & \textbf{Rank of Recom.}  \\
\midrule
A136 & 10 & 23, 26,31, 32, 34, 35, 37, 38, 47, 49 \\
A88 & 4 & 2, 19, 30, 50 \\
A816 & 10 & 3, 7, 8,9, 12, 13, 15, 44, 48 \\
A1437 & 1 & 21 \\
A138 & 6 & 5, 27, 28, 29, 36, 40\\
A128 & 1 &  24 \\
A295 & 7 & 1, 11, 14, 18, 22, 39, 41\\
A940 & 3 & 1, 16, 17\\
A976 & 8  & 6, 20, 25, 33, 42, 43, 45, 46 \\
\bottomrule
\end{tabular}
\caption{\textbf{Co-authorship Recommendations.} The rank links of discovered potential co-authorship for 9 sample researchers.}
\label{tab:rank2}
\end{table}

We additionally investigate the quality of recommendation of our model.     
A sample set of 9 researchers associated with the Linked Data and Information Retrieval communities~\cite{vahdati2018unveiling} are selected as the foundation for the experiments of the predicted recommendations. 
The top 20 recommendations per each researcher are filtered and integrated from the corresponding data,
out of 180 recommendations for all of the 9 selected researchers, 50 top predictions are filtered for a closer look. 
The results are validated by checking the research profile of the recommended researchers and the track history of co-authorship.

\section{Discussion and Future Work}
\label{sec:conclusion}
The aim of the present research was to develop a novel loss function to train the translational embedding model and examine it for graph completion of a real-world knowledge graph. 
This study has identified a successful application of TransESM, a modified version of TransE with soft marginal.
The results show the robustness of our model to deal with uncertainty in negative samples.
This reduces the negative effects of false negative samples on the computation of embeddings. 
Therefore, the performance of embedding model on knowledge graph completion task would be significantly improved. 
The proposed model has been applied on a scholarly knowledge graph.
The focus has been to discover example links between researchers with a potential of close scientific collaboration. 
The identified links have been proposed as collaboration recommendations and validated by looking into the profile of a list of selected researcher from the semantic web community.
As future work, we plan to apply the model on a broader scholarly knowledge graph and consider other different types of links for recommendations e.g, recommend events for researchers, recommend publications to be read or cited. 


%
\bibliographystyle{abbrv}
\bibliography{main.bbl}

\end{document}